\documentclass{article}
\usepackage{PRIMEarxiv}
\usepackage[utf8]{inputenc} 
\usepackage[T1]{fontenc}    
\usepackage{hyperref}       
\usepackage{url}            
\usepackage{booktabs}       
\usepackage{amsfonts}       
\usepackage{nicefrac}       
\usepackage{microtype}      
\usepackage{lipsum}
\usepackage{fancyhdr}       
\usepackage{graphicx}       
\graphicspath{{figures/}}    
\usepackage{placeins}
\usepackage{amsmath}
\usepackage[normalem]{ulem}
\usepackage{tabularray}
\title{HieraEdgeNet: A Multi-Scale Edge-Enhanced Framework for Automated Pollen Recognition
}

\author{
  Yuchong Long\textsuperscript{1,3}, \quad
  Wen Sun\textsuperscript{2,3}, \quad
  Ningxiao Sun\textsuperscript{1,3}, \quad
  Wenxiao Wang\textsuperscript{1,3}, \quad
  Chao Li\textsuperscript{1,3}, \quad
  Shan Yin\textsuperscript{1,3,*} \\
  \\
  \textsuperscript{1}School of Agriculture and Biology, Shanghai Jiao Tong University, 800 Dongchuan Rd., Shanghai 200240, China\\
  \textsuperscript{2}Shanghai Forestry Station, Hutai Rd. 1053, Shanghai, 200072, China\\
  \textsuperscript{3}Shanghai Yangtze River Delta Eco-Environmental Change and Management Observation and Research Station \\ (Shanghai Urban Ecosystem Research Station), Ministry of Science and Technology, \\ National Forestry and Grassland Administration, 800 Dongchuan Rd., Shanghai 200240, China\\
  \\
  \textsuperscript{*}Corresponding author \\
  \texttt{yinshan@sjtu.edu.cn} \\
  \texttt{ayinmostima@sjtu.edu.cn}
}

\begin{document}

\maketitle

\begin{abstract}
Automated pollen recognition is vital to paleoclimatology, biodiversity monitoring, and public health, yet conventional methods are hampered by inefficiency and subjectivity. Existing deep learning models often struggle to achieve the requisite localization accuracy for microscopic targets like pollen, which are characterized by their minute size, indistinct edges, and complex backgrounds. To overcome this limitation, we introduce HieraEdgeNet, a multi-scale edge-enhancement framework. The framework's core innovation is the introduction of three synergistic modules: the Hierarchical Edge Module (HEM), which explicitly extracts a multi-scale pyramid of edge features that corresponds to the semantic hierarchy at early network stages; the Synergistic Edge Fusion (SEF) module, for deeply fusing these edge priors with semantic information at each respective scale; and the Cross Stage Partial Omni-Kernel Module (CSPOKM), which maximally refines the most detail-rich feature layers using an Omni-Kernel operator—comprising anisotropic large-kernel convolutions and mixed-domain attention—all within a computationally efficient Cross-Stage Partial (CSP) framework. On a large-scale dataset comprising 120 pollen classes, HieraEdgeNet achieves a mean Average Precision (mAP@.5) of 0.9501, significantly outperforming state-of-the-art baseline models such as YOLOv12n and RT-DETR. Furthermore, qualitative analysis confirms that our approach generates feature representations that are more precisely focused on object boundaries. By systematically integrating edge information, HieraEdgeNet provides a robust and powerful solution for high-precision, high-efficiency automated detection of microscopic objects.
\end{abstract}


\section{Introduction}
Palynology, the scientific study of pollen grains and other micropaleontological entities, is of critical importance across diverse scientific domains, including paleoclimate reconstruction\cite{zargarPollenSporesProxies2025}, biodiversity monitoring\cite{braunImpactBiodiversityLoss2024,dwarakanathGlobalSurveyAddressing2024}, and allergenic disease research\cite{jumayevaPalynomorphologicalStudyAllergenic2023}. The type and concentration of airborne pollen are key indicators for assessing environmental quality and forecasting allergy seasons\cite{bergerDigitalHealthPatients2025,prodicInfluenceEnvironmentalPollution2025}. However, conventional pollen analysis relies heavily on manual microscopic enumeration by expert palynologists. This process is not only time-consuming and labor-intensive but is also prone to subjective error, severely limiting the throughput and timeliness of data acquisition\cite{gallardo-caballeroPrecisePollenGrain2019}. Consequently, the development of automated, high-throughput, and high-precision pollen identification technologies has become a pressing demand in the field.\par

The advent of computer vision and deep learning has presented new opportunities for automated pollen identification\cite{gargaPollenGrainClassification2024}. Object detection, a fundamental yet critical task in computer vision, aims to identify and localize instances of specific classes within an image\cite{lecunDeepLearning2015}. When applied to the analysis of microscopic images, particularly for pollen grain recognition, this task presents unique challenges. These challenges stem from the minute size of pollen grains, their embedding within complex backgrounds composed of dust and debris, their morphological diversity and tendency to aggregate, and their three-dimensional nature introduced by multi-focal plane imaging, which results in disparate appearances and sharpness for the same particle across different focal planes\cite{shamratPollenNetNovelDeep2024, gallardo-caballeroPrecisePollenGrain2019}. Early automated approaches primarily depended on traditional image processing techniques, such as color similarity transforms\cite{landsmeerDetectionPollenGrains2009}, active contour models\cite{nguyenImprovingPollenClassification2013}, or feature engineering based on shape and texture\cite{redondoPollenSegmentationFeature2015}. Although these methods achieved some success on specific datasets, they often required tedious manual feature design, and their robustness and generalization capabilities were often suboptimal when faced with real-world samples characterized by complex backgrounds, pollen aggregation, or morphological variability\cite{gallardo-caballeroPrecisePollenGrain2019,shamratPollenNetNovelDeep2024}. Deep learning, particularly the emergence of Convolutional Neural Networks (CNNs), offered a powerful new paradigm to address these issues\cite{lecunDeepLearning2015, gargaPollenGrainClassification2024}. By automatically learning hierarchical features from data, CNN-based object detection models have achieved revolutionary breakthroughs. In recent years, one-stage detectors, striking an exceptional balance between speed and accuracy, have rapidly become the mainstream approach for real-time object detection\cite{tianYOLOv12AttentioncentricRealtime2025}. These detectors frame the task as an end-to-end regression problem, thereby enabling real-time performance. In their architectural evolution, multi-scale feature representation is crucial for detecting objects of varying sizes. The Feature Pyramid Network (FPN), which constructs a feature pyramid via top-down pathways and lateral connections to effectively merge high-level semantic information with low-level spatial details, has become an indispensable component of modern detectors\cite{linFeaturePyramidNetworks2017}. Subsequent architectures combine these pyramid features with post-processing components like Non-Maximum Suppression (NMS) or Transformer-based decoders\cite{zhaoDetrsBeatYolos2024, carionEndtoendObjectDetection2020} for final detection.\par

The field of pollen identification and classification has also seen widespread application of these advances. Kubera et al.\cite{kuberaDetectionRecognitionPollen2022} applied the YOLOv5 model to detect three pollen types with high morphological similarity, demonstrating performance superior to other detectors like Faster R-CNN and RetinaNet. Tan et al.\cite{tanPollenDetectOpensourcePollen2022} also developed the PollenDetect system based on YOLOv5 for identifying pollen activity. These studies demonstrate the significant potential of deep learning models, especially the YOLO series, in automated palynology. However, the direct application of these general-purpose models to pollen identification is still confronted with distinct challenges. The minute size of pollen grains means they occupy a limited number of effective pixels in microscopic images\cite{gallardo-caballeroPrecisePollenGrain2019}. Consequently, relying solely on high-level semantic information is insufficient for high-precision recognition, rendering the effective extraction and utilization of boundary information paramount. Although some works have explored edge detection as an auxiliary task\cite{xieHolisticallynestedEdgeDetection2015}, how to systematically and synergistically fuse these low-level edge cues with high-level semantic features within the network remains a critical and unresolved issue.\par

To this end, we design and propose HieraEdgeNet, a multi-scale, edge-enhanced framework specifically engineered for high-precision automated pollen identification. Our architecture builds upon the foundational concepts previously discussed but introduces deep customizations and innovations to address the specific challenges of pollen analysis. The core innovation of HieraEdgeNet lies in three synergistic modules. First, we designed the Hierarchical Edge Module (HEM), which explicitly extracts a multi-scale edge feature pyramid in the early stages of the network, corresponding to the semantic hierarchy of the backbone. Subsequently, we introduce the Synergistic Edge Fusion (SEF) module, responsible for the deep, synergistic fusion of these scale-aware edge cues with semantic information from the backbone at each respective scale. Finally, to achieve ultimate feature refinement at the most detail-rich feature level, we adapt and incorporate the Cross Stage Partial Omni-Kernel Module (CSPOKM)\cite{weiYOLOCSPOKMEfficientSmall2025}, which integrates Omni-Kernel operators comprising large-scale, anisotropic convolutional kernels and mixed-domain attention mechanisms. Through these synergistic modules, HieraEdgeNet aims to significantly enhance the detection and localization accuracy of minute, indistinct targets, offering a robust solution for high-throughput, high-precision automated palynological analysis.

\section{Preliminaries and Related Work}

\subsection{Efficient Network Modules based on Cross Stage Partial Ideology}
To control computational costs while increasing network depth, efficient network architecture design is paramount. CSPNet (Cross Stage Partial Network) \cite{wangCSPNetNewBackbone2019} introduced an effective strategy that partitions the feature map entering a processing block along the channel dimension into two parts. One part undergoes deep transformation through the block, while the other is concatenated with the processed features via a short-circuit connection. This design reduces computational redundancy and enhances gradient propagation throughout the network, thereby improving model learning capacity and efficiency. In modern object detection architectures, the basic convolution operation is encapsulated within a standardized Conv block. This block conventionally combines a 2D convolution layer, a batch normalization layer, and an activation function in series. This Conv-BN-Act structure has been proven to effectively accelerate model convergence and improve training stability, while also providing a regularization effect \cite{ioffeBatchNormalizationAccelerating2015}. The C3k2 module, a core component for deep feature extraction in our architecture, is based on the design principles of CSPNet and adheres to the C2f paradigm \cite{sohanReviewYOLOv8Its2024}. This module is typically configured with standard Bottleneck blocks, each containing two convolution layers. The output of each processing unit is concatenated with the output of the previous stage, achieving a dense feature aggregation akin to that in DenseNet \cite{huangDenselyConnectedConvolutional2017}. Finally, the features from all branches are concatenated and passed through a 1x1 convolution for final feature fusion. This design not only minimizes computational redundancy but also facilitates network learning through enriched gradient paths. The Area-Attention C2f (A2C2f) module embeds a region-based attention mechanism (AAttn) into the efficient C2f structure, endowing the model with dynamic, context-aware capabilities while retaining the advantages of CNN's local inductive bias \cite{tianYOLOv12AttentioncentricRealtime2025}. The core of the A2C2f module is an ABlock, which comprises an AAttn module and a feed-forward network (MLP) with residual connections, structurally similar to a standard Encoder layer in a Vision Transformer (ViT) \cite{dosovitskiyImageWorth16x162020}. The AAttn module implements multi-head self-attention but can constrain the computation to specific regions of the feature map via an area parameter, enabling a trade-off between global attention and computational cost.\par
To enable the network to capture multi-scale contextual information from a single-scale feature map, the Spatial Pyramid Pooling - Fast (SPPF) module was introduced as an efficient variant \cite{jocherUltralyticsYolov5V32020,olorunsholaComparativeStudyYOLOv52023}. It employs a series of max-pooling layers to equivalently simulate the effect of parallel large-kernel pooling. An input feature map first undergoes channel reduction via a 1x1 convolution. Subsequently, this reduced feature map is passed serially through a max-pooling layer with a fixed kernel size (e.g., k=5) and a stride of 1, three consecutive times. The initial down-sampled feature is then concatenated along the channel dimension with the outputs from the three serial pooling operations. Since two consecutive kxk pooling operations (with stride 1) have a receptive field approximately equivalent to a single $(2k-1)x(2k-1)$ kernel, and three operations to a $(3k-2)x(3k-2)$ kernel, SPPF with k=5 serially applied three times can effectively capture receptive fields similar to those from parallel 5x5, 9x9, and 13x13 kernels in traditional SPP. Finally, the concatenated features are fused and their channels are adjusted by another 1x1 convolution. This serial design maintains multi-scale context awareness while significantly reducing computational complexity and improving inference speed.\par

\subsection{Feature Extraction and Complex Object Optimization}
The design of FPN enables modern object detectors to extract features at multiple scales, which is vital for detecting pollen grains of different sizes and locations. However, the limited resolution of microscopic images often results in objects that are minuscule with indistinct edges. Relying solely on standard multi-scale feature fusion is often insufficient for achieving precise detection in such contexts \cite{shamratPollenNetNovelDeep2024}. Consequently, enhanced strategies tailored to specific object characteristics are necessary. The identification and classification of pollen grains depend on both the boundary delimiting the grain from its environment and its internal features, making it crucial to enhance the model's representation capability for indistinct objects and edge details.\par
For specialized tasks, introducing prior knowledge as a strong inductive bias can significantly improve model performance and convergence speed. In the recognition of objects like pollen grains, precise boundary information is critical. The SobelConv module facilitates this by explicitly injecting first-order gradient information—i.e., edge features—into the network. A SobelConv module is essentially a fixed-weight 2D convolutional layer. Its kernel weights are initialized with the classic Sobel operators for computing horizontal and vertical image gradients \cite{sobel3x3IsotropicGradient1968}. After initialization, these weights are frozen to ensure operational consistency. In CNNs, standard downsampling operations like strided convolutions or pooling layers inevitably cause spatial information loss, which is particularly detrimental to the detection and structural analysis of fine-grained objects like pollen. To achieve lossless feature map downsampling, the Space-to-Depth Convolution (SPD-Conv) module was proposed \cite{sunkaraNoMoreStrided2022}. The core of this module is the "Space-to-Depth" transformation, also known as "Pixel Unshuffle." This operation rearranges a spatial block of 2x2 pixels into the channel dimension, effectively halving the spatial resolution while quadrupling the channel depth, thus achieving downsampling without information loss.\par
Traditional convolutional networks rely on a homogeneous paradigm of stacking small-kernel convolutions. To mine and transform input features from omnidirectional, multi-dimensional, and multi-domain perspectives, the Omni-Kernel Core operator was developed \cite{cuiOmnikernelNetworkImage2024,kvalnesOmnikernelOperatingSystem2014}. The core of this module simulates diverse receptive fields through a set of parallel and heterogeneous depth-wise separable convolutional kernels. Within the Omni-Kernel, Multi-Scale and Anisotropic Convolutions employ various kernel shapes in parallel \cite{dingScalingYourKernels2022}, including a large square kernel (e.g., 31x31) for capturing broad context and multiple anisotropic strip kernels (e.g., 1x31 and 31x1). These anisotropic kernels are particularly effective for perceiving structures with specific orientations, such as pollen contours or textures, thus providing an "omnidirectional" perceptual capability. The Omni-Kernel also integrates attention mechanisms from both the spatial and frequency domains. Frequency Channel Attention (FCA) first learns channel attention weights in the spatial domain via global average pooling and 1x1 convolutions, then applies these weights to the Fourier spectrum of the feature map \cite{qinFcanetFrequencyChannel2021}. This allows the model to dynamically adjust channel importance based on frequency composition. On the features processed by FCA, Spatial Channel Attention (SCA) further applies a classic Squeeze-and-Excitation type of channel attention for secondary calibration in the spatial domain \cite{huSqueezeandexcitationNetworks2018}. The Frequency Gating Module (FGM) acts as a dynamic frequency-domain filter, selectively enhancing or suppressing specific frequency components via a learned gating mechanism, further improving the model's adaptability to complex patterns \cite{raoGlobalFilterNetworks2021}. The Omni-Kernel design is intended to enhance the expressive power for complex pollen morphologies, improve discrimination of subtle inter-class differences, and increase localization accuracy amidst complex backgrounds and ambiguous boundaries.\par

\subsection{Datasets for Pollen Recognition}
The development of robust, automated pollen recognition systems is critically dependent on large-scale, high-quality, and meticulously annotated datasets. Publicly available resources for pollen analysis, summarized in Table \ref{datasets}, can be broadly categorized into two main types: those comprising pre-segmented, single-grain images designed for classification tasks, and those providing fully annotated microscopic scenes for object detection. These datasets collectively offer a diverse range of pollen types, imaging conditions, and annotation complexities, providing a comprehensive foundation for benchmarking. This study leverages a selection of these public datasets to evaluate the performance of the HieraEdgeNet architecture, aiming to advance the accuracy, robustness, and usability of automated pollen recognition.

\begin{table}[ht]
\centering
\caption{Datasets for Pollen Recognition}
\label{datasets} 
\begin{tblr}{
  width = \linewidth,
  colspec = {Q[331]Q[138]Q[473]},
  cells = {c},
  hline{1-2,9} = {-}{2pt},
}
Dataset                             & Abbreviation & Source                                                         \\
POLLEN73S                           & POLLEN73S    & \cite{astolfiPOLLEN73SImageDataset2020}      \\
POLEN23E                            & POLEN23E     & LifeCLEF 2023                                                  \\
POLLEN20L                           & POLLEN20L    & \cite{khanzhinaCombatingDataIncompetence2022} \\
ABCPollenOD                         & ABCPollenOD  & \cite{kuberaDetectionRecognitionPollen2022}  \\
Cretan Pollen Dataset v1 (CPD-1)  & Cretan       & \cite{galan2021cretan}                       \\
Monospecific Mediterranean Pollen & Monospecific & \cite{gregor2023monospecific}               \\
Pollen Video Library                & PollenVideo  & \cite{caoAutomatedPollenDetection2020}      \\
\end{tblr}
\end{table}

\section{Methodology}
\begin{figure}[htbp]
    \centering
    \includegraphics[width=0.6\linewidth]{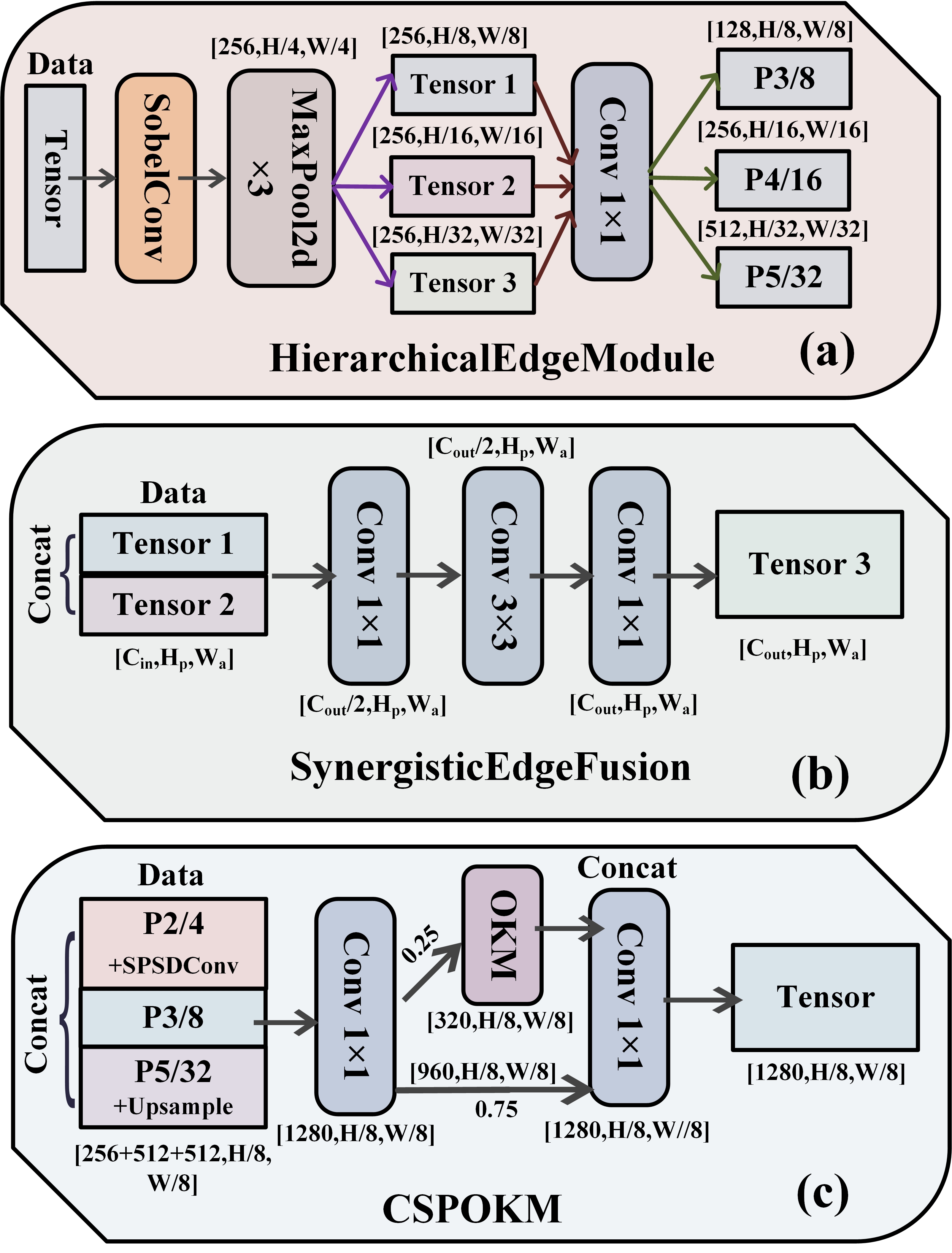}
    \caption{The three core modules of HieraEdgeNet: (a) The HEM explicitly extracts and organizes multi-scale edge information from input features. (b) The SEF module effectively fuses scale-specific edge features, extracted by the HEM, with semantic features from the corresponding scale of the backbone network. (c) The CSPOKM introduces robust multi-scale and omnidirectional receptive field feature extraction, integrating diverse large-kernel, anisotropic depth-wise separable convolutions alongside spatial and frequency domain attention mechanisms}
    \label{Keyarc}
\end{figure}

To address the challenges inherent in detecting pollen grains—characterized by their minute size, limited effective pixel count, and the critical reliance of precise localization on the effective differentiation of target edges from complex environmental backgrounds—we introduce HieraEdgeNet. This novel architecture is engineered to substantially enhance the recognition and localization accuracy of small objects, with a particular focus on pollen grains, by employing hierarchical and synergistic perception of edge information integrated with multi-scale feature fusion. The core innovation of HieraEdgeNet lies in the introduction of three pivotal structural modules: the Hierarchical Edge Module (HEM), the Synergistic Edge Fusion (SEF) module, and the Cross Stage Partial Omni-Kernel Module (CSPOKM)\cite{weiYOLOCSPOKMEfficientSmall2025}. These modules operate synergistically to form a network highly sensitive to edge information and endowed with robust feature representation capabilities. The architectural details of these three core modules are illustrated in Fig. \ref{Keyarc}.

\FloatBarrier
\FloatBarrier

\subsection{Hierarchical Edge Module}
HEM is engineered to extract and construct a pyramid of multi-scale edge information from input feature maps, thereby explicitly capturing fine-grained image details and salient edge characteristics. We denote feature maps from stage $k$ of a network backbone as $Pk$; these are typically characterized by a spatial resolution downscaled by a factor of $2^k$ relative to the input image (e.g., $P2$ features are at $1/4$ resolution, $P3$ at $1/8$, and so forth). The HEM initially employs a fixed Sobel operator, implemented as a convolutional layer with non-learnable weights (SobelConv), to independently compute gradients for each channel of an input feature map, $X_{\text{in}}$. This $X_{\text{in}}$ is assumed to be from an early backbone stage, such as $P2$ (i.e., $1/4$ resolution, e.g., with 256 channels), yielding an initial edge response map, $X_{edge}^{(0)}$. This map $X_{edge}^{(0)}$ preserves the spatial resolution and channel dimensionality of $X_{\text{in}}$. Subsequently, max-pooling layers (MaxPool2d) are iteratively applied to progressively downsample $X_{edge}^{(0)}$, generating edge maps at successively lower resolutions: $X_{edge}^{(1)}$ (at $1/8$ resolution relative to the original image input) and $X_{edge}^{(2)}$ (at $1/16$ resolution). Finally, these hierarchically generated edge maps ($X_{edge}^{(0)}$, $X_{edge}^{(1)}$, and $X_{edge}^{(2)}$) are each processed by an independent 1x1 convolution. These convolutions adjust the channel dimensions to produce the multi-scale edge features $E_{P3}$ (128 channels, derived from $X_{edge}^{(0)}$ at $1/4$ res.), $E_{P4}$ (256 channels, from $X_{edge}^{(1)}$ at $1/8$ res.), and $E_{P5}$ (512 channels, from $X_{edge}^{(2)}$ at $1/16$ res.). These edge features ($E_{P3}$, $E_{P4}$, $E_{P5}$) are thus generated at spatial resolutions of $1/4$, $1/8$, and $1/16$ respectively, and are designated for subsequent fusion with semantic features from backbone stages $P3$, $P4$, and $P5$ (which are typically at $1/8$, $1/16$, and $1/32$ resolutions, respectively).\par
Let $X_{\text{in}}$ denote an input feature map. The HEM performs the following operations:
\begin{equation}
\begin{aligned}
X_{edge}^{(0)} &= \text{SobelConv}(X_{in}) \\
X_{edge}^{(1)} &= \text{MaxPool}(X_{edge}^{(0)}) \\
X_{edge}^{(2)} &= \text{MaxPool}(X_{edge}^{(1)}) \\
E_{P3} &= \mathcal{F}_{1\times1}^{(C_{P3})}(X_{edge}^{(0)}) \\
E_{P4} &= \mathcal{F}_{1\times1}^{(C_{P4})}(X_{edge}^{(1)}) \\
E_{P5} &= \mathcal{F}_{1\times1}^{(C_{P5})}(X_{edge}^{(2)})
\end{aligned}
\end{equation}
where $\text{SobelConv}(\cdot)$ denotes the Sobel edge extraction implemented via fixed-weight convolution, $\text{MaxPool}(\cdot)$ represents max pooling with a kernel size of 2 and a stride of 2, and $\mathcal{F}_{1\times1}^{(C_{\text{target}})}(\cdot)$ signifies a 1x1 convolutional block that adjusts the feature map to $C_{\text{target}}$ output channels. $E_{P3}, E_{P4}, \text{and } E_{P5}$ are the resulting multi-scale edge feature maps, with $C_{P3}, C_{P4}, \text{and } C_{P5}$ being their respective channel dimensions (128, 256, and 512, as specified above).\par
The HEM furnishes the backbone network with explicit edge cues at multiple abstraction levels, designed for fusion with corresponding semantic features. This mechanism significantly aids the model in achieving more precise localization of object boundaries and in comprehending fine-grained structural details.\par

\subsection{Synergistic Edge Fusion}
SEF is engineered to integrate core semantic information, crucial for pollen classification, with salient edge cues essential for precise localization. SEF operates by concurrently processing semantic features extracted at three primary spatial scales within the network backbone (P3/8, P4/16, and P5/32, corresponding to feature maps downsampled by factors of 8, 16, and 32, respectively). These semantic features are then fused with the corresponding scale-specific edge features generated by the HEM.\par
Let $X_{main}$ denote the input semantic features and $X_{edge}$ represent the input edge features from the corresponding scale. The symbol $\bigoplus$ signifies concatenation along the channel dimension. The operations within the SEF module are defined as follows:
\begin{equation}
\begin{aligned}
X_{cat} &= X_{main} \bigoplus X_{edge} \\
X_{mid1} &= \mathcal{F}_{1\times1}^{(C_{out}/2)}(X_{cat}) \\
X_{mid2} &= \mathcal{F}_{3\times3}^{(C_{out}/2)}(X_{mid1}) \\
Y_{SEF} &= \mathcal{F}_{1\times1}^{(C_{out})}(X_{mid2})
\end{aligned}
\end{equation}
where $Y_{SEF}$ is the fused output feature map of the SEF module. Here, $C_{out}$ specifies the number of output channels for the SEF module. The application of SEF modules at successive stages of the network, each operating on features of decreasing spatial resolution due to backbone downsampling, facilitates a continuous and hierarchical process of feature fusion. This mechanism allows SEF to effectively meld semantic and edge information, thereby substantially enhancing the model's proficiency in both the classification and precise localization of pollen grains.

\subsection{Cross Stage Partial Omni-Kernel Module}
The primary objective of the CSPOKM is to enhance the flexibility and expressive power of feature extraction through the strategic incorporation of Omni-Kernel capabilities within a cross-stage, partial processing framework. This module is designed to effectively capture multi-scale features by leveraging inputs from different network stages. Specifically, CSPOKM introduces a learnable ensemble of convolutional kernels via the Omni-Kernel, which are characterized by their spatial sharing and channel-wise distinctions, enabling the model to flexibly combine and adapt feature representations at various levels of abstraction. Within HieraEdgeNet, the CSPOKM embeds the Omni-Kernel—a sophisticated operator comprising diverse large-scale, multi-directional depth-wise convolutions, augmented by attention and gating mechanisms such as SCA, FCA, and a FGM—into a CSP architecture. This configuration is employed for feature refinement within a critical path of the detection head, specifically at the P3 level, aiming to enhance the detail and spatial precision of features by integrating multi-scale contextual information.\par
The operational flow begins by concatenating features from disparate pathways: $X_{P2S}$ (derived from P2/4 features processed by an SPD-Conv layer, designed to reduce spatial dimensions while preserving feature information), $X_{P3B}$ (the fused backbone features at the P3/8 level), and $X_{P5U}$ (obtained by upsampling P5/32 features). Let the concatenated input features be $X_{in} = X_{P2S} \bigoplus X_{P3B} \bigoplus X_{P5U}$. The subsequent operations within CSPOKM are detailed as follows:

 \begin{equation}
 \begin{aligned}
X_{feat} &= \mathcal{F}_{1\times1}^{(C_1)}(X_{in}) \\
(X_{okm\_in}, X_{skip}) &= \text{Split}_{e}(X_{feat}) \\
X_{okm\_out} &= \mathcal{M}_{OKM}(X_{okm\_in}) \\
Y_{CSPOKM} &= \mathcal{F}_{1\times1}^{(C_2)}(X_{okm\_out} \bigoplus X_{skip})
\end{aligned}
\end{equation}

Here, $\bigoplus$ denotes concatenation along the channel dimension. The $\text{Split}{e}(\cdot)$ function partitions the feature channels into two segments based on a ratio $e$ (empirically set to $e=0.25$ in our architecture) and $1-e$. The segment $X_{okm_in}$ is processed by the Omni-Kernel module, denoted as $\mathcal{M}_{OKM}(\cdot)$, yielding $X{okm\_out}$. This output is then concatenated with the skipped connection $X_{skip}$ and further processed by a $1 \times 1$ convolutional block to produce the final output $Y_{CSPOKM}$. Typically, to maintain channel consistency for feature refinement at a specific level, $C_1$ and $C_2$ are set equal to a target channel dimension $C_{target}$, ensuring that the module's input and output channel counts are concordant.\par
This sophisticated processing, particularly at critical feature levels rich in detail such as P3, endows the network with a powerful feature learning capability. It enables the capture of complex patterns and long-range dependencies that are challenging for conventional convolutional layers. Concurrently, the CSP architecture ensures computational efficiency. Such characteristics are particularly advantageous for improving performance in scenarios involving small objects or requiring fine-grained distinctions between classes.

\subsection{HieraEdgeNet}
\begin{figure}[htbp]
    \centering
    \includegraphics[width=0.9\linewidth]{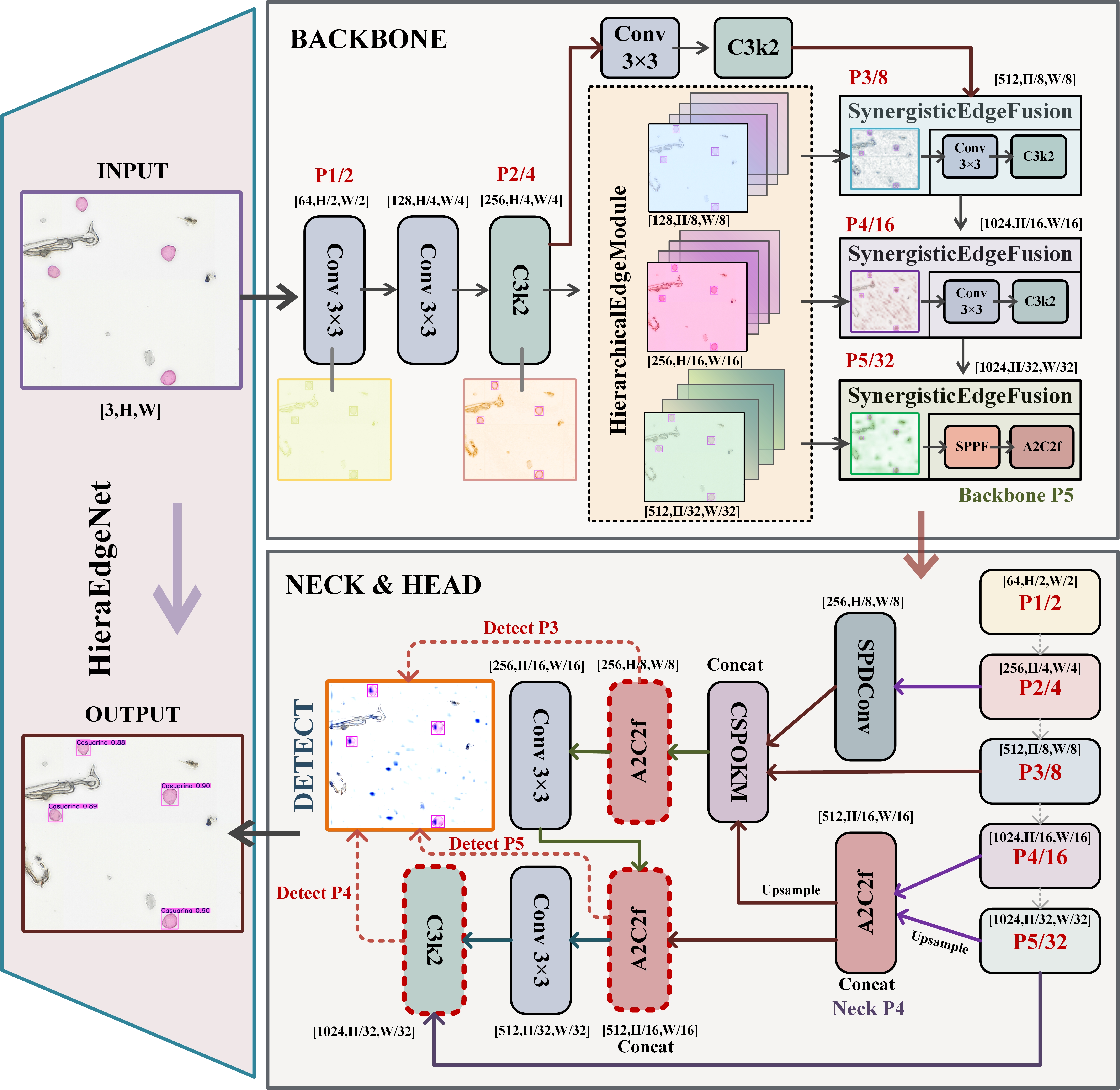}
    \caption{The complete architecture of HieraEdgeNet, featuring enhancements for small objects, complex background information, and blurred edges.}
    \label{Net}
\end{figure}

HieraEdgeNet is a deep convolutional neural network meticulously designed for object detection tasks where targets, such as pollen grains visualized in microscope slides, necessitate localization based on boundary information and classification using internal semantic information. Its core characteristic is the explicit extraction and fusion of multi-scale edge information, which is synergistically combined with the semantic features from the backbone network. This approach aims to enhance detection accuracy, particularly in scenarios involving complex background information and blurred target edges encountered across diverse microscopy environments. The central design philosophy involves introducing a parallel HEM within the backbone network. The multi-scale edge features extracted by the HEM in the early stages of the network are subsequently fed into the SEF modules alongside the semantic features from the backbone. The neck network adopts principles from the Path Aggregation Network (PANet) to further amalgamate multi-scale features. Crucially, within this stage, the CSPOKM is deployed to specifically address the challenges of detecting small and indistinct targets. Finally, detection heads operating on three distinct feature map scales, enriched with fused multi-scale features, perform the object predictions. The comprehensive architecture of HieraEdgeNet is depicted in Fig. \ref{Net}.

\FloatBarrier
\FloatBarrier

In the initial downsampling phase of the network, two sequential 3x3 Conv modules, each with a stride of 2, are employed for rapid downsampling, generating P1/2 (64 channels, H/2, W/2) and P2/4 (128 channels, H/4, W/4) feature maps, respectively. Subsequently, main feature extraction and parallel edge information processing are conducted. The P2/4 feature map is processed by a C3k2 module, comprising two repeating units, which outputs a 256-channel feature map, thereby enhancing the semantic representation at the P2/4 level. Concurrently, the HEM operates in parallel, directly processing the output of the P2/4 stage. Following this, the backbone network extracts semantic features at three critical scales—P3/8, P4/16, and P5/32—and fuses them with the corresponding scale-specific edge features. The SEF module concatenates the extracted P3/8 edge features (128 channels) with the P3/8 semantic features from Layer 8 of the backbone (512 channels). This concatenated feature set is then deeply fused through a sequence of a 1x1 Conv (for channel fusion and dimensionality reduction to 256 channels), a 3x3 Conv (for spatial feature extraction, 256 channels), and another 1x1 Conv (for channel adjustment to 512 channels), yielding a 512-channel P3/8 fused feature map. Similarly, the extracted P4/16 edge features (256 channels) are fused with P4/16 semantic features (512 channels), resulting in a 512-channel P4/16 fused feature map. Further, the P5/32 edge features (512 channels) are combined with P5/32 semantic features (1024 channels) to produce a 1024-channel P5/32 fused feature map. A total of three such SEF operations are sequentially employed within the backbone. Feature enhancement continues at the terminus of the backbone. An SPPF (Spatial Pyramid Pooling - Fast) module is applied to the P5/32 fused features, utilizing serial max-pooling operations (with equivalent kernel sizes of 5, 9, and 13) to further augment multi-scale contextual awareness, outputting 1024 channels. Subsequently, four repeating units of the A2C2f module, employing ABlock (Area Block, a regional attention mechanism, where area=1 signifies global regional attention), perform deep optimization and contextual modeling on the P5/32 features, also outputting 1024 channels. This constitutes the highest-level semantic feature output from the backbone, designated Backbone P5.\par
The neck network adopts a PANet-like architecture, integrating features from different levels of the backbone through top-down and bottom-up pathway aggregation to generate a more discriminative multi-scale feature pyramid for subsequent detection. The top-down feature enhancement path initiates with P5 to P4 level feature fusion: the highest-level semantic features from the backbone (Backbone P5) are upsampled via nearest-neighbor interpolation and concatenated with the P4/16 fused features from the backbone. This aggregated feature map is then refined by an A2C2f module (internally employing a C3k structure), producing the Neck P4 feature map (512 channels). Next, P4 to P3 level feature fusion and innovation occur: the Neck P4 features are upsampled and concatenated with the P3/8 fused features from the backbone. Notably, to bolster small object detection capabilities, high-resolution detail supplementary features, derived from the early P2/4 features of the backbone, are introduced; these are downsampled to the P3/8 resolution and their channel dimensions adjusted using an SPDConv module, designed for lossless downsampling. These three feature streams (upsampled Neck P4, backbone P3/8 fused features, and SPDConv-processed P2/4 features) are concatenated and input into the core CSPOKM module. This CSPOKM module, leveraging its internal Omni-Kernel components, deeply optimizes these multi-source P3 features. Finally, further refinement through an A2C2f module generates the Detect P3 feature map (256 channels). The bottom-up feature localization path commences with P3 to P4 level feature fusion: the Detect P3 features are downsampled by a Conv module with a stride of 2 and concatenated with the Neck P4 features generated in the top-down path. This aggregated feature map is then refined by an A2C2f module to yield the Detect P4 feature map (512 channels) for detection. Lastly, P4 to P5 level feature fusion is performed: the Detect P4 features are downsampled by a Conv module with a stride of 2 and concatenated with the Backbone P5 features. This combination is then optimized by a C3k2 module (internally employing a C3k structure) to generate the Detect P5 feature map (1024 channels) for detection.\par
Through this bidirectional pathway aggregation and the optimization provided by specific modules, the neck network ultimately outputs three feature maps at different scales (Detect P3, Detect P4, and Detect P5) to the detection head. These feature maps effectively balance high-level semantic information with precise spatial localization capabilities.

\subsection{Detection Head and Loss Function}
The detection architecture directly employs a decoupled head. During the forward propagation phase, for each detection level $l$, the input feature map $X_l \in \mathbb{R}^{B \times C_l \times H_l \times W_l}$ is processed independently by the regression and classification heads. The regression head outputs $P_{reg}^{(l)} = \mathcal{F}_{reg}^{(l)}(X_l) \in \mathbb{R}^{B \times 4R \times H_l \times W_l}$, and the classification head outputs $P_{cls}^{(l)} = \mathcal{F}_{cls}^{(l)}(X_l) \in \mathbb{R}^{B \times N_c \times H_l \times W_l}$. During inference, predictions from all levels are concatenated and decoded. For each detection layer $l \in {1, \dots, L}$ and at each spatial location $(i,j)$, let the regression prediction be $P_{reg}^{(l)}[i,j] \in \mathbb{R}^{4R}$ and the class prediction be $P_{cls}^{(l)}[i,j] \in \mathbb{R}^{N_c}$. For bounding box decoding, $P_{reg}^{(l)}[i,j]$ is partitioned into four components $d_t, d_b, d_l, d_r \in \mathbb{R}^{R}$, corresponding to the distance distributions for the top, bottom, left, and right sides of the bounding box, respectively. After processing via the mechanism associated with Distribution Focal Loss (DFL), these yield scalar distances $\delta_t = \mathcal{D}(d_t)$, $\delta_b = \mathcal{D}(d_b)$, $\delta_l = \mathcal{D}(d_l)$, and $\delta_r = \mathcal{D}(d_r)$, where $\mathcal{D}(\cdot)$ represents the operation deriving a scalar distance from the learned distribution. The class probabilities are defined as $p^{(l)}[i,j] = \sigma(P_{cls}^{(l)}[i,j]) \in [0,1]^{N_c}$, where $\sigma$ denotes the sigmoid function.
\par
Addressing the imbalance in pollen samples, this study adopts the Focal Loss ($\mathcal{L}_{FL}$) as the classification loss function. Focal Loss introduces a modulating factor $(1-p_t)^\gamma$ and an optional balancing factor $\alpha_t$, $\gamma \ge 0$ is the focusing parameter. If $p_k$ is the model's predicted probability for the $k$-th class and $\hat{p}_k$ is the corresponding ground truth label (typically 0 or 1), the Focal Loss for the $k$-th class is:
 \begin{equation}
\mathcal{L}_{FL}^{(k)} = -\alpha_k (1-p_{t,k})^\gamma \log(p_{t,k})
\end{equation}
where $p_{t,k} = p_k$ if $\hat{p}_k=1$, and $p_{t,k} = 1-p_k$ if $\hat{p}_k=0$.
The total loss for object detection, $\mathcal{L}_{DetTotal}$, is a weighted sum of the regression and classification losses, with weights $\lambda_{reg}$ and $\lambda_{cls}$ respectively:
\begin{equation}
\mathcal{L}_{DetTotal} = \lambda_{reg} \mathcal{L}_{reg} + \lambda_{cls} \mathcal{L}_{cls}
\end{equation}
Here, $\mathcal{L}_{cls}$ is the sum of Focal Losses computed over all positive samples and selected negative samples. The bounding box regression loss ($\mathcal{L}_{reg}$) employs Distribution Focal Loss (DFL) to learn the distribution of distances from the anchor/reference point to the four sides of the bounding box. This is typically used in conjunction with an IoU-based loss, such as CIoU, GIoU, or SIoU loss. The classification loss ($\mathcal{L}_{cls}$), as previously mentioned, is calculated using Focal Loss.

\section{Experiment}

\subsection{Dataset Preparation}
\begin{figure}[htbp]
    \centering
    \includegraphics[width=0.9\linewidth]{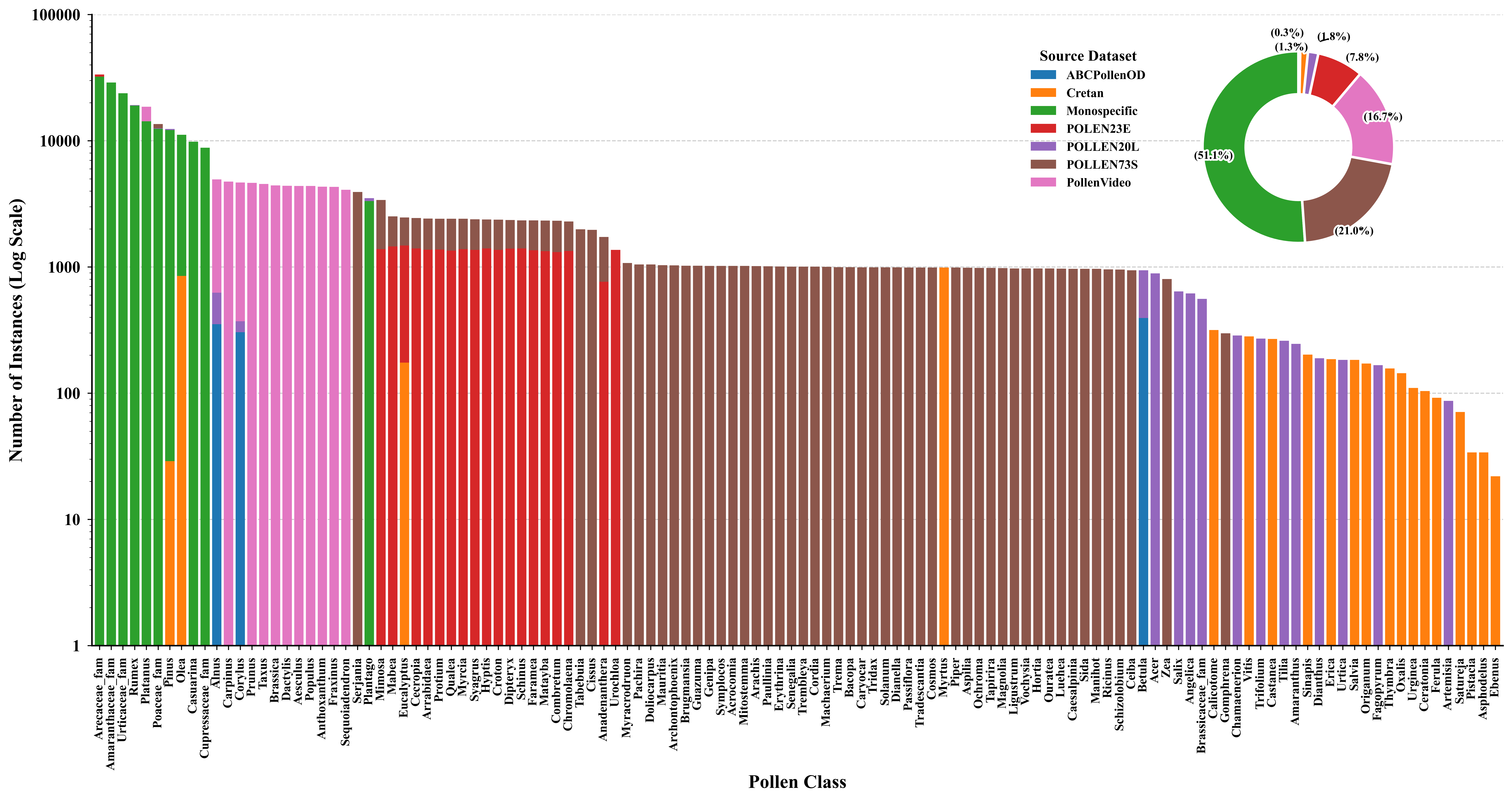}
    \caption{The distribution of different pollen types in the dataset, highlighting the diversity and scale of the training samples.}
    \label{dataset_distribution}
\end{figure}

To facilitate the development and rigorous evaluation of HieraEdgeNet, we constructed a large-scale pollen detection dataset comprising 120 distinct pollen categories. The dataset was aggregated from two primary sources: (1) several existing, manually annotated pollen detection datasets, and (2) a novel data synthesis pipeline designed to convert a vast collection of single-grain classification images into detection samples with precise bounding box annotations. This synthesis strategy involves programmatically embedding individual pollen grain images onto authentic microscopy backgrounds, thereby substantially augmenting both the scale and diversity of the training data. The final dataset exhibits a characteristic long-tailed class distribution (Fig. \ref{dataset_distribution}), which closely mimics real-world scenarios. This feature is critical for training a high-performance model that is robust in practical applications. For model training, the dataset was partitioned into a training set (80\%) and a validation set (20\%), maintaining a proportional class distribution based on the number of instances per category. The model was trained for 500 epochs. During training, we employed a suite of data augmentation techniques—including mosaic augmentation, Gaussian blur, and color space shifting—to enhance the model's adaptability and generalization to complex, unseen target domains.\par

\FloatBarrier
\FloatBarrier

\subsection{Quantitative Evaluation and Benchmarking}

To systematically evaluate the performance of our proposed HieraEdgeNet architecture, we conducted a comprehensive set of experiments, benchmarking it against several state-of-the-art (SOTA) real-time object detectors. These baseline models include the CNN-based YOLOv11n\cite{khanamYOLOv11OverviewKey2024} and YOLOv12n\cite{tianYOLOv12AttentioncentricRealtime2025}, as well as the Transformer-based RT-DETR-R18 and RT-DETR-R50\cite{jiaDetrsHybridMatching2023}. Furthermore, we conducted two ablation studies on HieraEdgeNet. First, we created a hybrid model by integrating its backbone with the detection head of RT-DETR (designated HieraEdgeNet-RT-DETR) to validate the versatility and feature extraction capability of its backbone. Second, we applied the Layer-wise Automated Model Pruning (LAMP) structured pruning technique\cite{yangLAMPSLayerwisedMixedprecisionandsparsity2024} to create a compressed variant (HieraEdgeNet-LAMP), aiming to explore its deployment potential in resource-constrained scenarios. Our evaluation employs standard COCO metrics, including mAP@.5, mAP@.75, and the primary metric, mAP@.5:.95, alongside a comprehensive assessment of model parameters, computational complexity (GFLOPs), and inference speed (FPS).\par
The quantitative results, presented in Table \ref{accuracy1}, clearly demonstrate that our proposed HieraEdgeNet achieves the highest overall detection accuracy among the evaluated models. Its mAP@.5:.95 score of 0.8444 surpasses the advanced, similarly-sized models YOLOv12n and YOLOv11n by 1.29 and 2.09 percentage points, respectively. This significant accuracy gain provides strong evidence for the effectiveness of our designed HierarchicalEdgeModule, SynergisticEdgeFusion, and CSPOmni-Kernel modules in enhancing feature representation. Although this precision enhancement is accompanied by a moderate increase in computational cost (GFLOPs of 14.0 versus 6.7 for YOLOv12n), this trade-off is highly valuable given the stringent accuracy requirements of the pollen recognition task.\par
The advantages of HieraEdgeNet are even more pronounced when compared to the Transformer-based RT-DETR models. In terms of accuracy, HieraEdgeNet's mAP@.5:.95 outperforms RT-DETR-R18 by 3.7 percentage points and RT-DETR-R50 by 5.48 percentage points. In terms of efficiency, HieraEdgeNet's model size (3.88M parameters / 14 GFLOPs) is substantially lower than that of RT-DETR-R18 (20.03M / 57.5 GFLOPs) and RT-DETR-R50 (42.18M / 126.2 GFLOPs). This indicates that our architecture, through deep structural innovation within a CNN framework, achieves a superior accuracy-efficiency balance compared to leading Transformer-based detectors.When our backbone is integrated with the RT-DETR decoder, the resulting hybrid model (HieraEdgeNet-RT-DETR) achieves an mAP@.5:.95 of 0.8492—the highest of all models tested. This result compellingly demonstrates the high quality and generalizability of the features produced by our backbone, capable of providing robust support for diverse detector heads and surpassing the original RT-DETR-R18 and R50. Finally, addressing the practical need for lightweight models, our pruned HieraEdgeNet-LAMP variant achieves a remarkable mAP@.5:.95 of 0.8363 with significantly reduced parameters and computation. This accuracy not only exceeds that of the RT-DETR series but also both YOLOv11n and YOLOv12n, while maintaining a highly competitive inference speed (403.24 FPS). This showcases the excellent compressibility of the HieraEdgeNet architecture, its superior accuracy-efficiency curve, and its significant potential for practical applications.\par

\begin{table}[ht]
\centering
\caption{Quantitative performance evaluation of HieraEdgeNet and its variants against state-of-the-art detectors. The comparison includes metrics for model complexity (Parameters, GFLOPs, Model Size), inference speed (FPS), and detection accuracy (mAP@.5, mAP@.75, mAP@.5:.95) on the pollen dataset. The better results in each accuracy column are highlighted in bold.}
\label{accuracy1} 
\begin{tblr}{
  width = \linewidth,
  colspec = {Q[156]Q[131]Q[94]Q[177]Q[83]Q[88]Q[88]Q[119]},
  cells = {c},
  hline{1-2,9} = {-}{2pt},
}
Model        & Parameters & GFLOPs & Model Size (MB) & FPS    & mAP@.5           & mAP@.75           & mAP@.5:.95        \\
HieraEdgeNet & 3,882,080  & 14     & 7.8             & 361.17 & \textbf{0.9501} & \textbf{0.9304} & \textbf{0.8444} \\
YOLOv12n      & 2,628,224  & 6.7    & 5.4             & 534.44 & 0.9478          & 0.9225          & 0.8315          \\
YOLOv11n      & 2,653,648  & 6.7    & 5.4             & 577.77 & 0.9486          & 0.9215          & 0.8235          \\
HieraEdgeNet-RT-DETR & 73,096,064 & 211.7 & 140.4 & 49.39  & 0.9444 & \textbf{0.9262} & \textbf{0.8492} \\
HieraEdgeNet-LAMP   & 2,688,579  & 11.6  & 5.6   & 403.24 & 0.9394 & 0.9189          & \textbf{0.8363} \\
RT-DETR-R18   & 20,025,584 & 57.5   & 38.9            & 177.37 & 0.9201          & 0.8975          & 0.8074          \\
RT-DETR-R50   & 42,181,284 & 126.2  & 164.7           & 73.61  & 0.9169          & 0.8886          & 0.7896          
\end{tblr}
\end{table}

\begin{figure}[htbp]
    \centering
    \includegraphics[width=0.9\linewidth]{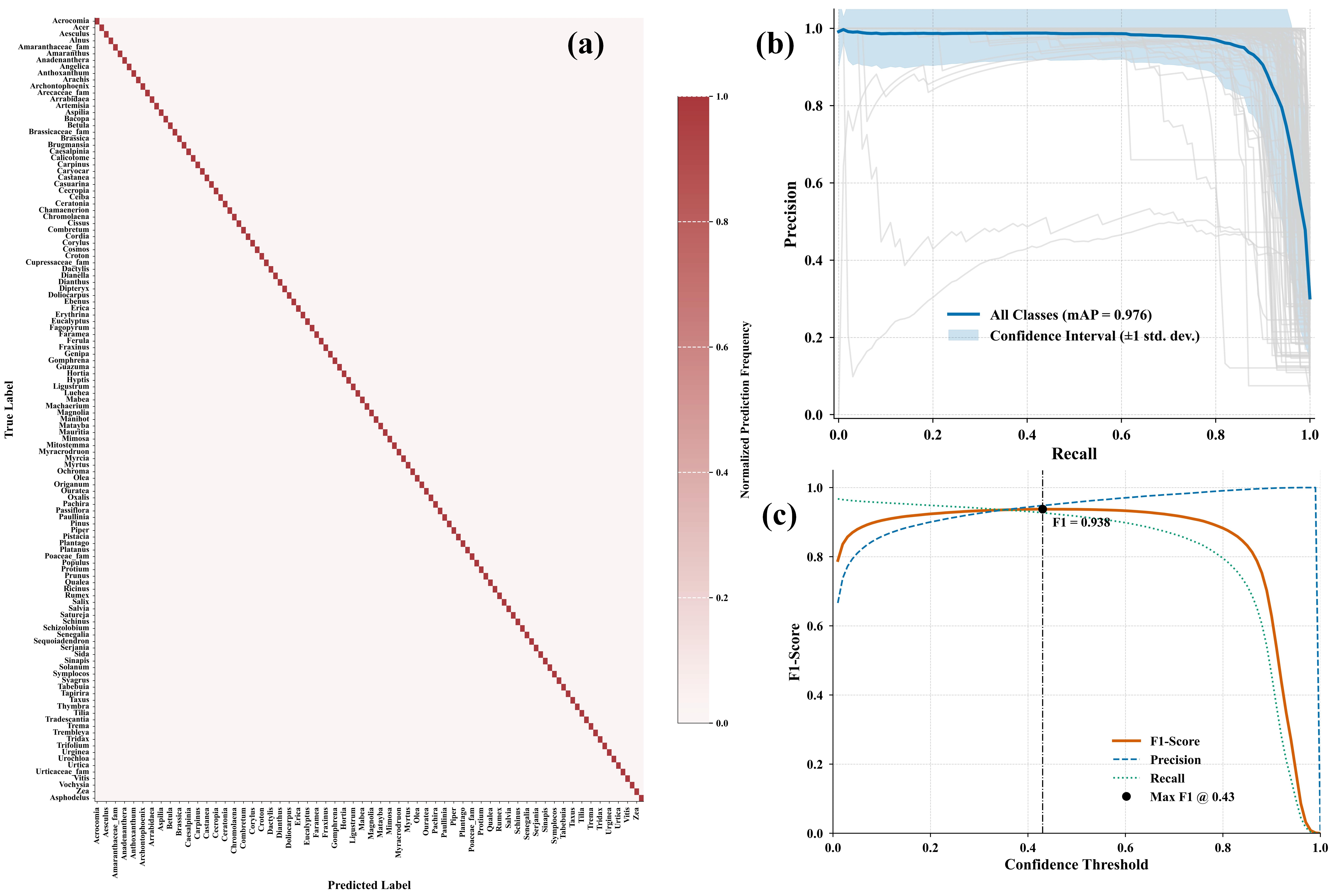}
    \caption{Detailed performance validation of the proposed HieraEdgeNet model. (a) The confusion matrix for all pollen classes, demonstrating high classification accuracy and minimal inter-class confusion. (b) The Precision-Recall (P-R) curve, with the mean Average Precision (mAP) value over all classes indicated. (c) The F1 score curve as a function of the confidence threshold, highlighting the peak F1 score and the corresponding optimal threshold.}
    \label{valmodel}
\end{figure}

The detection performance of HieraEdgeNet was further analyzed, as depicted in Figure \ref{valmodel}. The confusion matrix (Fig. \ref{valmodel}a) exhibits a distinct and highly concentrated diagonal, indicating extremely high classification accuracy across the 46 pollen classes with minimal inter-class confusion. The Precision-Recall (P-R) curve (Fig. \ref{valmodel}b) further corroborates the model's exceptional performance, achieving a mean Average Precision (mAP) of 0.976 over all classes. The broad area under the curve demonstrates that the model maintains high precision across various recall levels. The F1-score curve (Fig. \ref{valmodel}c) illustrates the model's comprehensive performance at different confidence thresholds, reaching a peak F1-score of 0.938 at an optimal confidence threshold of 0.43. This provides a reliable basis for selecting the optimal operating point for the model in practical deployments.\par

\FloatBarrier
\FloatBarrier

In summary, both the horizontal comparisons against state-of-the-art CNN and Transformer detectors and the vertical analyses of our model variants consistently affirm that the HieraEdgeNet architecture delivers superior precision and efficiency for automated pollen recognition through its deep mining and fusion of edge information and multi-scale features.\par

\subsection{Visual Analysis of Enhanced Edge Perception}

\begin{figure}[htbp]
    \centering
    \includegraphics[width=0.9\linewidth]{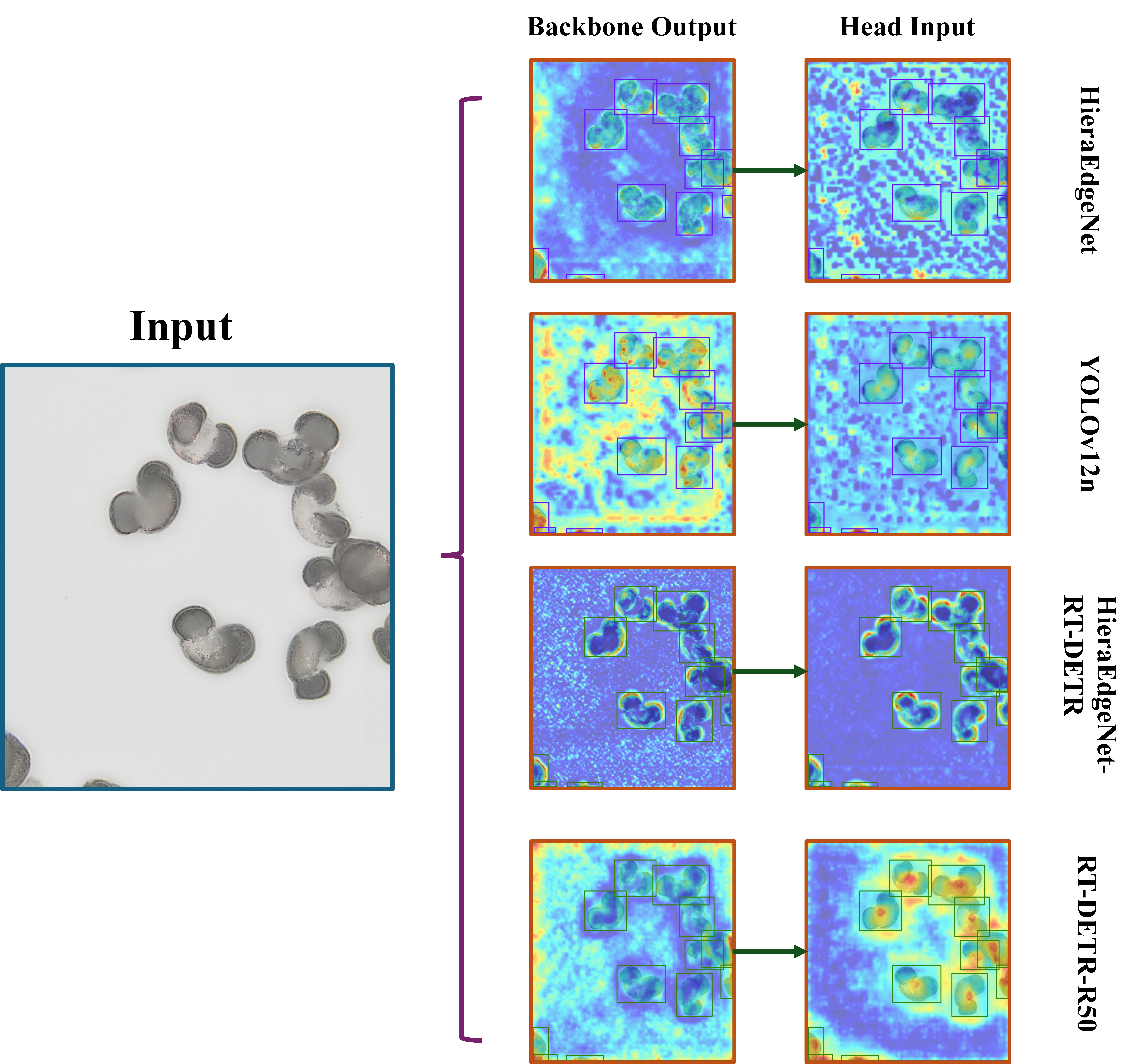}
    \caption{Visual analysis of HieraEdgeNet's enhanced edge perception. Grad-CAM heatmaps compare the activation focus of HieraEdgeNet against baseline models at two key stages: the backbone output (top row) and the detection head input (bottom row). HieraEdgeNet shows a distinctly sharper and more accurate focus on the pollen grain edges, which is crucial for its superior performance.}
    \label{edge}
\end{figure}

Figure \ref{edge} presents a comparative visualization using Gradient-weighted Class Activation Mapping (Grad-CAM)\cite{selvarajuGradCAMVisualExplanations2020} heatmaps for four models—HieraEdgeNet, YOLOv12n, HieraEdgeNet-RT-DETR, and RT-DETR-R50—at both the backbone's output layer and the detection head's input layer. These heatmaps visually render the model's attention intensity across the input image, with a specific focus on the edges and structural details of pollen grains.\par
Our analysis reveals that HieraEdgeNet generates more refined and sharply focused responses along the object boundaries, reflecting the HEM's effective capture of edge information. This advantage is further amplified at the input of the detection head, where the representation of target edges is enhanced. In contrast, models like YOLOv12n and RT-DETR-R50 do not exhibit a comparable level of edge sensitivity, particularly at the detection head's input layer, where their edge responses appear more diffuse. This suggests a deficiency in their explicit utilization of edge information, which may consequently limit their localization accuracy. In summary, the performance superiority of HieraEdgeNet is rooted in its ability to generate more precise and less noisy feature maps—a capability that is crucial for precise object localization within complex microscopic environments.\par

\FloatBarrier
\FloatBarrier

\section{Conclusion}
In this work, we present HieraEdgeNet, a deep learning framework featuring multi-scale edge enhancement, specifically engineered for automated pollen recognition. To overcome the challenges of minute target size, indistinct edges, and complex backgrounds characteristic of pollen grains, HieraEdgeNet successfully integrates explicit, multi-scale edge priors with deep semantic information through a highly efficient and synergistic fusion mechanism. This is accomplished via three pivotal innovations: the HEM, the SEF module, and the CSPOKM. The HEM systematically extracts an edge feature pyramid aligned with the backbone network's hierarchy; the SEF facilitates deep interaction between edge and semantic features at each corresponding scale; and the CSPOKM, deployed at the critical P3 feature level, performs meticulous optimization of fine-grained details using its internal Omni-Kernel operators, which comprise large-scale anisotropic convolutions and mixed-domain attention mechanisms. The efficacy of HieraEdgeNet is empirically substantiated by extensive experiments. In benchmark comparisons against state-of-the-art real-time object detectors, including YOLOv12n and RT-DETR, HieraEdgeNet achieved superior performance, significantly surpassing all baseline models on the critical mAP@.5:.95 metric. Furthermore, the HieraEdgeNet backbone demonstrated remarkable generalization capabilities when integrated with the RT-DETR decoder. Moreover, a structurally pruned version of the model retained high accuracy while exhibiting outstanding potential for practical deployment. Qualitative analysis using Grad-CAM visualizations further confirmed that our architecture generates feature responses that are more precisely focused and localized to object boundaries. Collectively, these findings demonstrate that by deeply mining and integrating edge information, HieraEdgeNet provides a robust and reliable solution for high-precision, high-efficiency automated pollen recognition.\par
Despite its demonstrated high accuracy, HieraEdgeNet has limitations, primarily stemming from the substantial computational cost introduced by advanced components such as the Omni-Kernel. Additionally, its inherently two-dimensional design precludes the direct utilization of three-dimensional (Z-stack) microscopy data, and its performance may be challenged by domain shifts between training data and real-world samples. Consequently, our future work will focus on three key directions: (1) optimizing computational efficiency through techniques like model compression and acceleration; (2) extending the edge-enhancement framework to three dimensions to process volumetric data; and (3) employing domain adaptation methods to improve the model's generalization and robustness across varied acquisition conditions.

\section*{Additional information}
The  dataset used for training and evaluating the models in this study is publicly available on Kaggle at \href{https://www.kaggle.com/datasets/ayinven/hieraedgenetintegratesdatasets}{https://www.kaggle.com/datasets/ayinven/hieraedgenetintegratesdatasets}. The models are deposited on the Hugging Face Hub and can be accessed at \href{https://huggingface.co/datasets/AyinMostima/HieraEdgeNetintegratesdatasets}{https://huggingface.co/datasets/AyinMostima/HieraEdgeNetintegratesdatasets}. The source code for the complete pollen identification framework, PalynoKit, is available in the GitHub repository at \href{https://github.com/AyinMostima/PalynoKit}{https://github.com/AyinMostima/PalynoKit}.

\section*{Acknowledgments}
The computations in this paper were performed on the $\pi$ 2.0, supported by the Center for High Performance Computing at Shanghai Jiao Tong University. We sincerely thank the center for their technical support and resources.

\bibliographystyle{unsrt}  
\bibliography{references}

\end{document}